# Multimodal Meaning Representation for Generic Dialogue Systems Architectures

Frédéric Landragin, Alexandre Denis, Annalisa Ricci, Laurent Romary

LORIA – UMR 7503
Campus scientifique
B.P. 239
F-54506 Vandœuvre-lès-Nancy Cedex
{landragi, denis, ricci, romary}@loria.fr

**Abstract**
An unified language for the communicative acts between agents is essential for the design of multi-agents architectures. Whatever the type of interaction (linguistic, multimodal, including particular aspects such as force feedback), whatever the type of application (command dialogue, request dialogue, database querying), the concepts are common and we need a generic meta-model. In order to tend towards task-independent systems, we need to clarify the modules parameterization procedures. In this paper, we focus on the characteristics of a meta-model designed to represent meaning in linguistic and multimodal applications. This meta-model is called MMIL for MultiModal Interface Language, and has first been specified in the framework of the IST MIAMM European project. What we want to test here is how relevant is MMIL for a completely different context (a different task, a different interaction type, a different linguistic domain). We detail the exploitation of MMIL in the framework of the IST OZONE European project, and we draw the conclusions on the role of MMIL in the parameterization of task-independent dialogue managers.

## Introduction

The specification of a language that represents both the form and the content of linguistic resources is an important task in the design of dialogue systems architectures. The more spontaneous and constraints-free is the natural language dialogue, the more complex are the form and the content of resources. When the dialogue is multimodal, so when a gesture capture device is associated to the microphone with which the user interacts with the system, the language shall combine the capability of handling complex structures in the language resources with the generality and the flexibility required for operating as communication interface between the various modules. In a multimodal system, the main and classical modules are the following: speech recognizer, gesture recognizer, semantic analyzer, multimodal fusion, action planner, multimodal fission, speech synthesizer, visual feedback producer.

The use of a representation language common for all communicative acts offers several advantages in terms of generality and parameterization. For instance, exchanges between all the previously mentioned modules will be represented using the same format and the same content description, and the particular application for which the system is instanced will parameterize the action planner using the same type of resource.

In this paper, we present our experience in designing the MMIL (MultiModal Interface Language) language in the framework of the IST MIAMM European project (see also Kumar & Romary, 2002), and we describe the procedure while re-using it for the IST OZONE European project. In particular, we describe the MMIL specifications for the two demonstrators that were implemented during these projects, and the adaptations required for the management of new features. Among this new features: the management of salience, the status of secondary events in an user utterance, and the status of speech acts. On the basis of this experience, we draw some conclusions on the design of application-independent dialogue systems, and we discuss how MMIL is the object of a possible future standardization for the representation of multimodal semantic content.

## MMIL specifications in MIAMM

The specification of MMIL in MIAMM copes the definition of a language able to uniformly represent semantic content in a multimodal context, so to capture:
- linguistic, gestural, and graphical events,
- both dialogue acts and dialogue act's contents.

Past or existing initiatives are often limited to specific modalities, while the aim of MMIL is to provide a meta-model for semantic representation free from any modality constraint.

### MMIL compared to other languages

M3L (Multimodal Markup Language) was specified for the SmartKom project for the representation of information that flows between the various processing components (speech recognition, gesture recognition, face interpretation, media fusion, presentation planning, etc.). In particular, M3L represents all information about segmentation, synchronization and confidences in processing results. Its main strong point is in its large coverage. But, contrary to MMIL, there is no meta-model behind M3L, and its XML syntax is not so flexible.

EMMA (Extensible MultiModal Annotation Markup Language) aims to represent information automatically extracted from a user's input by an interpretation component. It's a technical report from the W3C Multimodal Interaction working group, which main purpose is to develop specifications to enable access to the Web using multimodal interaction. This orientation makes this language very engineering-oriented.

MPML (Multimodal Presentation Markup Language) was and is still designed for multimodal presentation using interactive life-like agents. Its purpose is to provide a means to write attractive presentations easily (Tsutsui *et al.*, 2000). It illustrates the interest of description language for multimodal generation and not only interpretation. But it's very procedural.

MURML (Multimodal Utterance Representation Markup Language) bridges between the planning and the animation tasks in the production of multimodal utterances of an anthropomorphic agent (Kranstedt *et al.*, 2002). Its main strong points are in the description of gestural behaviors and in the will of merging representations from interpretation and generation.

ULF (Ahn *et al.*, 1995) and UMRS (Copestake, 1995; Egg, 1998) are examples of interesting approaches that focus on linguistic meaning representation using logical forms. But they have not yet been extended to multi-modality.

To summarize, we can identify three main approaches when specifying a multimodal content description language. First, the formalism-oriented approaches that focus on formal properties. ULF and UMRS belong to them, with an integration of complex logical forms. Second, the engineering-oriented approaches, that, face to the complexity of particular systems, focus on their particular needs. This is typically the case of M3L and MPML. The problem with such approaches is in their poor reusability. Third, the ontology-oriented approaches that constitute a compromise solution. Focusing on the concepts and on the management of under-specification, MMIL is an example of such an approach. When designing MMIL and for its future improvements, we aim to include the maximum of categories, and to tend to a format that covers the maximum of phenomena, from purely lexical aspects to pragmatic speech acts in dialogue.

**A short description of MMIL**

The MMIL meta-model abstracts different levels of dialogue information (phone, word, phrase, utterance) by means of a flat ontology which identifies shared concepts and constraints. The definition layer of the ontology includes two kinds of entities: events and participants. Events are objects associated to the temporal level, while participants are static entities acting upon or being affected by the events. Dependencies between entities are represented as typed relations linking structural nodes. Contrary to other semantic information models, the MMIL meta-model does not include relations, which are perceived as qualifying descriptors defining anchors among entities. As the other information units of the MMIL model (e.g., morpho-syntactic, domain, annotation descriptors), relations act in the information architecture as a set of descriptors (data categories) that formally describe the specification constraints. The data categories, expressed in an RDF format compatible with ISO 11179-3, give the necessary openness to the design of the semantic structures, so to cope with the potential flexibility of the model.

## Testing MMIL for OZONE

To test the language's application-independence and the generality of its meta-model, we migrated MMIL from MIAMM to the OZONE project. OZONE differs from MIAMM in the following respects:
- the interaction mode (tactile vs. haptic gesture),
- the application domain (train vs. music),
- the task type (requests vs. database querying).

The context is then very different and some problems arise. After a short presentation of the architecture, we discuss here three issues: (a) the distinction between attention and salience for graphical and gestural events; (b) the status of secondary events; (c) the status of speech acts.

**Architecture and use of MMIL in OZONE**

MMIL is used for the representation of all information communicated between the various agents of the architecture. These agents are the following ones:
- tactile gesture interpretation,
- speech interpretation, that is based on a speech recognition module and whose role is to parse the utterance,
- multimodal fusion, that integrates the results of the two previous modules with the dialogue history,
- action planner, where the system decides how to react to an utterance,
- modality adviser, that (a) tests the usability of speech and visual modalities for the feedback of information, and (b) chooses the most relevant one considering the context,
- application(s), each ones including a client and a server,
- response generation, including the following modules: multimodal fission, speech synthesizer, visual feedback producer.

**The management of salience**

From a theoretical point of view (Rousselet *et al.*, 2003), attention and salience differ on how they influence perception. Attention originates in the user (top-down), whereas salience originates in the context (bottom-up). In dialogue systems, attention can be identified when objects are selected or mentioned by the user (low-term memory). MIAMM manages such a principle, but it does not address the issue of salience scores, introduced in MMIL to handle the representations of OZONE's complex visual scenes.

Figure 1 shows such a scene, including some train stations and some ways to reach them by train. A gesture is circling the station of Meudon (with the highest salience) and four ways to reach Meudon (with lower saliences). The MMIL corresponding file is shown in Figure 2.

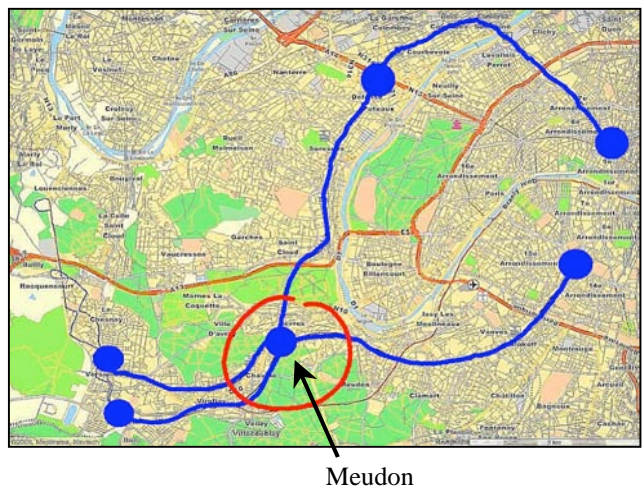

Meudon

Figure 1: Screenshot from OZONE demonstrator

```xml
<mmil:mmilComponent xmlns:laf="http://www.tc37sc4.org/laf"
xmlns:mmil="http://www.miamm.org/mmil">
<mmil:event id="e0">
   <mmil:evtType>VTState</mmil:evtType>
   <mmil:dialogueAct>inform</mmil:dialogueAct>
   <mmil:tempSpan mmil:endPoint="Tue Oct 28 13:19:05 CET
    2003" mmil:startPoint="Tue Oct 28 13:19:04 CET 2003"/>
</mmil:event>
<mmil:event id="e1">
   <mmil:evtType>report</mmil:evtType>
   <mmil:actionStatus>performed</mmil:actionStatus>
</mmil:event>
<mmil:participant id="p0">
   <mmil:MMILId>MEUDON</mmil:MMILId>
   <mmil:salience>26</mmil:salience>
   <mmil:attentionStatus>inSelection</mmil:attentionStatus>
</mmil:participant>
<mmil:relation laf:source="e1" laf:target="e0"type="propContent"/>
<mmil:relation laf:source="p0" laf:target="e1" type="description"/>
</mmil:mmilComponent>
```

Figure 2: MMIL file for the interpretation of a gesture

This example illustrates the representation and the exploitation of salience scores when interpreting an ambiguous gesture. The complete disambiguation will be possible when confronting the result of the gesture interpretation to the result of the linguistic interpretation.

## The status of secondary events

Complex sentences need MMIL to cope with improved mechanisms for representing the status of secondary events. In the management of linguistic events, the MIAMM's basic assumption is to systematically attach the propositional content's event to the sole verb of the sentence (e.g., "I want pop or rock"). In OZONE a sentence can involve more than one verb and, depending on the case, more than one event. The role of each event within the semantic structure (main event or secondary event) is often related to the syntactic approach. For instance, we can compare the following sentences:
1. "I want to go to Paris"
2. "I must go to Paris"

It is easy to see the different relevance of the "go" event: secondary event in (1), main event in (2). The MMIL representation that is obtained for (1) is given in Figure 3.

```xml
<mmil:mmilComponent xmlns:laf="http://www.tc37sc4.org/laf"
xmlns:mmil="http://www.miamm.org/mmil">
<mmil:event id="e0">
   <mmil:speaker>user</mmil:speaker>
   <mmil:evtType>speak</mmil:evtType>
   <mmil:addressee>system</mmil:addressee>
   <mmil:dialogueAct>request</mmil:dialogueAct>
   <mmil:spokenLanguage>en</mmil:spokenLanguage>
</mmil:event>
<mmil:event id="e1">
   <mmil:evtType>want</mmil:evtType>
   <mmil:mode>indicative</mmil:mode>
   <mmil:tense>present</mmil:tense>
</mmil:event>
<mmil:event id="e2">
   <mmil:evtType>go</mmil:evtType>
   <mmil:mode>indicative</mmil:mode>
   <mmil:tense>present</mmil:tense>
</mmil:event>
<mmil:participant id="p0">
   <mmil:lex>i</mmil:lex>
   <mmil:objType>PERSON</mmil:objType>
   <mmil:refType>1PPDeixis</mmil:refType>
</mmil:participant>
<mmil:participant id="p1">
   <mmil:lex>paris</mmil:lex>
   <mmil:objType>PLACE</mmil:objType>
</mmil:participant>
<mmil:relation laf:source="e1" laf:target="e0"type="propContent"/>
<mmil:relation laf:source="p0" laf:target="e1" type="subject"/>
<mmil:relation laf:source="e2" laf:target="e1"type="object"/>
<mmil:relation laf:source="p1" laf:target="e2" type="destination"/>
</mmil:mmilComponent>
```

Figure 3: MMIL file for a linguistic representation

## The status of speech acts

The speech act of an utterance reflects the way the user wants the system to react. Speech acts are linked to pragmatics of dialogue and MMIL shall represent them, since the language is intended to cover the different levels of dialogue information. Following Relevance Theory (Sperber & Wilson, 1995), we consider as the three fundamental speech acts saying, telling and asking. With P the propositional form of the utterance, we have:
- saying that P,
- telling the hearer to P,
- asking Wh-P (what, who, where, why, etc.).

MIAMM involves the representation of the following speech acts: open, close, inform, request, accept, and reject. But the MIAMM application is only working with orders, and the resultant dialogue act is always a request (for instance, "play rap from the 90's"), whatever the form of the utterance, saying ("I want rap from the 90's"), telling ("Please play rap from the 90's") or asking ("Can you play rap from the 90's?"). OZONE differentiates the three forms. Saying and asking share in MMIL the same structure, but different polarities affect the entities. For instance, the sentence "How can I go to Paris?" is represented the same as "I can go to Paris by X", where an interrogative polarity is put on the entity representing X (see Figure 4). Concerning indirect speech acts, e.g., an assertion that hides an order, they do not address MMIL and have to be resolved in the dialogue modules.

```xml
<mmil:mmilComponent xmlns:laf="http://www.tc37sc4.org/laf"
xmlns:mmil="http://www.miamm.org/mmil">
<mmil:participant id="0">
   <mmil:lex>i</mmil:lex>
   <mmil:objType>PERSON</mmil:objType>
   <mmil:refType>1PPDeixis</mmil:refType>
</mmil:participant>
<mmil:participant id="1">
   <mmil:lex>paris</mmil:lex>
   <mmil:objType>PLACE</mmil:objType>
   <mmil:mmilId>Paris</mmil:mmilId>
</mmil:participant>
<mmil:participant id="2">
   <mmil:question>how</mmil:question>
</mmil:participant>
<mmil:event id="3">
   <mmil:evtType>go</mmil:evtType>
   <mmil:mode>indicative</mmil:mode>
   <mmil:tense>present</mmil:tense>
   <mmil:modal>can</mmil:modal>
</mmil:event>
<mmil:event id="4">
   <mmil:speaker>user</mmil:speaker>
   <mmil:evtType>speak</mmil:evtType><mmil:addressee>
```

```xml
    system</mmil:addressee>
   <mmil:dialogueAct>request</mmil:dialogueAct>
   <mmil:spokenLanguage>en</mmil:spokenLanguage>
</mmil:event>
<mmil:relation laf:source="3" laf:target="4" type="propContent"/>
<mmil:relation laf:source="0" laf:target="3" type="subject"/>
<mmil:relation laf:source="1" laf:target="3" type="destination"/>
<mmil:relation laf:source="2" laf:target="3" type="mean"/>
</mmil:mmilComponent>
```

Figure 4: MMIL file for a question

## Conclusion

The work we have conducted on the definition and tuning of the MMIL language in the two European projects described in this paper can be seen as a kind of experiment to identify precise requirements on what a general framework for multimodal content representation. Those requirements should obviously go beyond what has been described in (Bunt & Romary, 2002), in order to identify classes of applications which bear enough features to be covered by one single model (or at least meta-model in the sense of Ide & Romary, 2002). Indeed, it may not be likely that the kind of representations needed for such applications as information extraction, named entity recognition, reference annotation (see Salmon-Alt & Romary, 2004), or the annotation of temporal structure will be based on exactly the same underlying structures.

Still, it seems necessary that those various types of models do share a common semantics for any sub-structure they would share and even more for any elementary descriptor they would use (e.g., a certain dialogue act /inform/, or discourse relation /elaboration/, a temporal relation /overlap/, or an elementary role in relation to an event /agent/). Such a goal obviously requires that there is some kind of consensus on providing some shared definition of such concepts, as well as an international infrastructure to submit, select and disseminate those descriptors. The first aspect is one of the topics which has been considered as underlying the activity of the ACL/SIGSEM working group on multimodal semantic content representation and is being pursued through a series of meetings that have taken place since November 2002.

The second aspect is the core of a standardizing effort in ISO committee TC 37 to deploy an on-line data category registry intended to cover a wide variety of descriptors (also known as *data categories* in the TC 37 terminology) identified in existing representation or annotation practices.

In this context, we would like to see MMIL as one instance of such a descriptive and modelling activity which would nicely fit the needs of multimodal dialogue system when conveying meaning from one component to another. If it is the case, we could also contemplate using MMIL — or a dialect thereof — for such tasks as the evaluation of dialogue systems. This is what we currently explore in the context of the French project Media dedicated to the contextual evaluation of dialogue manager modules.

## Acknowledgements

This work was supported by the IST-2000-29487 MIAMM EC project and the IST-2000-30026 OZONE EC project (see References).